\documentclass[11pt]{article}
\usepackage{apacite}
\usepackage{graphicx}
\usepackage[margin=0.75in]{geometry}
\usepackage{amssymb}
\usepackage{natbib}
\usepackage{pdflscape}
\usepackage{caption}
\usepackage{tabularx}
\usepackage{pdflscape}
\usepackage{appendix}

\begin{document}
\title{Towards Improving Predictive Risk Modelling for New Zealand's Child Welfare System Using Clustering Methods\thanks{This work was supported by Auckland University of Technology, School of Engineering, Computer and Mathematical Sciences Doctoral Scholarship, Auckland, New Zealand.}\footnotemark[1]~~\thanks{Access to the data used in this study was provided by Stats NZ under conditions designed to give effect to the security and confidentiality provisions of the Data and Statistics Act 2022. The results in this paper are not official statistics. They have been created for research purposes from the Integrated Data Infrastructure (IDI) which is carefully managed by Stats NZ. The opinions, findings, recommendations, and conclusions expressed in this paper are those of the authors, not Stats NZ. For more information about the IDI please
visit https://www.stats.govt.nz/integrated-data/.}\footnotemark[2]}
\author{%
  Sahar Barmomanesh\textsuperscript{1}\thanks{Corresponding author (e-mail: sahar.barmomanesh@ aut.ac.nz).}, Victor Miranda-Soberanis   \textsuperscript{1}\\[0.25ex]
  \begin{tabular}{@{}c@{}}
    \textsuperscript{1}Department of Mathematical Sciences, Auckland University of Technology, \\6 St Paul Street, Auckland City, Auckland, 1010, New Zealand 
  \end{tabular}%
}

\date{} 
\maketitle

\begin{abstract}
The combination of clinical judgment and predictive risk models crucially assist social workers to segregate children at risk of maltreatment and decide when authorities should intervene. Predictive risk modelling to address this matter has been initiated by several governmental welfare authorities world-wide involving administrative data and machine learning algorithms. While previous studies have investigated risk factors relating to child maltreatment, several gaps remain as to understanding how such risk factors interact and whether predictive risk models perform differently for children with different features. 
By integrating Principal Component Analysis and K-Means clustering, this paper presents initial findings of our work on the identification of such features as well as their potential effect on current risk-modelling frameworks.
This approach allows examining existent, unidentified yet, clusters of New Zealand (NZ) children reported with 
\textit{care and protection concerns}, as well as
to analyze their inner structure, and evaluate the performance of prediction models trained cluster--wise. 
We aim to discover the extent of clustering degree required as an early step in the development of predictive risk models for child maltreatment and so enhance the accuracy of such models intended for use by child protection authorities. The results from testing LASSO logistic regression models trained on identified clusters revealed no significant difference in their performance. The models, however, performed slightly better for two clusters including younger children. our results
suggest that separate models might need to be developed for children of certain age to gain additional control over the error rates and to improve model accuracy. While results are promising, more evidence is needed to draw definitive conclusions, and further investigation is necessary.\newline
\textbf{\textit{Keywords:}} Child welfare, Administrative data, Predictive risk models, Clustering analysis,
Machine learning
\end{abstract}
\section{Introduction}
Child welfare authorities across the nations are involved in daily high-risk projects towards risk assessment of children at risk of harm from their caregivers, and appropriate measures as to when authorities should intervene. These risk-prediction projects depend on decision-makers who often manage overwhelming caseloads under high stress, with limited time, and insufficient information \citep{glaberson2019coding}. This is often an important source of risk-prediction bias slipping into critical human-driven decisions. Existing research emphasizes the limitations and biases of clinical judgment and argues that the conjunction of clinical judgment and a predictive tool can contribute to more efficient decision-making \citep{cuccaro2017risk,wilson2015predictive,vaithianathan2013children,rea2017report}.

A growing number of child welfare authorities around the world are developing and deploying predictive tools that utilize government data and a machine learning algorithm to predict the likelihood of child maltreatment. Within the United States (US), for instance, predictive tools are currently being tested and used by the child protection agencies. The three most well-recognized predictive tools currently in use by child protection agencies in the US are the Eckerd Rapid Safety Feedback tool (ERSF), developed in Florida \citep{eckerd2016report}, the Allegheny Family Screening Tool (AFST), developed by and used in Allegheny County, Pennsylvania \citep{vaithianathan2019allegheny}, and the Douglas County Decision Aid model (DCDA) developed specifically for Douglas County in Colorado \citep{vaithianathan2019implementing}. As part of its early intervention prevention strategy, the NZ government has also considered using predictive tools within its child welfare system as an assistance tool in decision-making \citep{ministry2012white}. Predictive risk models that would utilize linked administrative data and machine learning algorithms to identify children who are at high priority for preventive services. The machine learning algorithm incorporates measures of the child, young person, and their family’s characteristics efficiently and consistently to generate a single measure of overall risk. A limited number of predictive risk models were developed to assess the technical feasibility and predictive validity of the NZ government’s proposal \citep{wilson2015predictive,vaithianathan2013children}. For instance, \citet{vaithianathan2013children} focused on young children in NZ social welfare system and developed a model to predict the risk of substantiated finding of maltreatment before the age of 5 for every child entering the public benefit system before the age of two. Meanwhile, \citet{wilson2015predictive} examined the technical feasibility and predictive validity of the proposal with a population-wide predictive risk model. The authors focused on a model that would predict whether a new-born child will experience maltreatment by age five. The findings from both studies suggest that predictive risk models based on administrative data have the potential to identify children whose preventive needs are high. However, predictive tools should be considered as complementary to professional judgment rather than as a substitute and should not be considered the sole method for identifying high risk children. 

Furthermore, after a number of ethical reviews \citep{dare2013predictive,blank2015ethical} and feasibility studies \citep{ministry2014feasibility} on the use of predictive risk modeling within NZ child welfare system, the NZ government commissioned a project to explore the use of predictive risk modeling within their child protection agency’s intake decision-making. \citet{rea2017report} provided a comprehensive report on this project. According to  \citet{rea2017report}, the goal of the project was to use the information available in administrative data to predict potential concerns for child maltreatment at the time the contact center receives a care and protection notification. The model was developed using 70 percent of unique children and young people under 15 years of age who were notified to the agency in 2013 and the testing data set consisted of all the notifications not used in the training data set and covered the period January 2011 to Jun 2014. Four machine learning algorithms, including logistic regression, decision tree, random Forest, and gradient boosting were tested. Despite having a similar Area Under the ROC Curve (AUC)\footnote{\textbf{Area Under the ROC curve (AUC)} is a performance measurement for the classification problems at various threshold settings. ROC is a probability curve, and AUC represents the degree or measure of separability. It tells how much the model is capable of distinguishing between classes. Higher the AUC, the better the model is at predicting 0s as 0s and 1s as 1s.} of 0.75 to that of the boosting and random forest models, logistic regression was chosen as the model of choice owing to its simplicity and ease of interpretation. The assessment of the model based on the percentage of children in each group with a true concern for child maltreatment within 24 months revealed that the model is reasonably accurate in discriminating between high risk and low risk children and young people. However, discrimination is more difficult for group of children in the middle risk score deciles. Despite this limitation, the project's findings suggest that predictive tools can enhance care and protection intake decisions for NZ children, young people, and their families. However, there are still concerns regarding the accuracy and fairness of such models \citep{gillingham2017predictive,keddell2019algorithmic}. Consequently, further work is required to mitigate these concerns before NZ government can proceed to an implementation phase.

Making decisions that affect children's lives is often a critical task and that predicting a child future risk of maltreatment at the time of notification is complicated. A predictive risk model is subject to errors and consequently may incorrectly identify as low risk some children who go on to experience abuse or neglect (False Negative (FN)) as well as identify as high-risk some children who do not (False Positive (FP)) \citep {de2014another}. Clearly, these two types of errors may result in different harms: a FP may result in an unnecessary intervention or even the separation of a family. At the same time, a FN could lead the agency to fail to intervene when it should have . In child welfare, both classification errors are of concern. FNs can be dangerous to the child, while FPs can result in poor targeting of agency resources \citep{dare2013predictive,blank2015ethical}. Therefore, careful attention must be paid to the accuracy of these models. 

The current study is motivated by the need for further research on improving the accuracy and fairness of predictive risk models developed for potential use within NZ child welfare system. Due to confidentiality obligations, we were unable to obtain access to the data used in the NZ government-commissioned project. As a result, we created our own research data set from a combination of literate and academic studies that identified risk factors of abuse and maltreatment, primarily those described in \citep{rea2017report}. More details on the data creation process are provided in Section 2.1. For the purpose of enhancing the accuracy of these models, we investigate the possibility of using Clustering Analysis (CA) methods as an early step in the development of predictive risk models for use within the child protection agency’s intake decision-making. By Using CA methods we expect children to be assigned to groups (clusters) so that children within each group are like one another. In contrast to the classification problem where each observation belongs to one of several groups, and the aim is to predict the group to which an observation belongs, CA seeks to discover the number and composition of the groups by identifying discrete, potentially hidden, groups of children. As part of our approach, we analyze the inner structure of these clusters to identify certain characteristics of children that may have a substantial impact on the error rate of these models and so be able to determine whether these errors can be reduced by training separate models for these subgroups of children. The specific objectives are \textbf{1)} to create a research data set including the outcome variable and predictor variables, \textbf{2)} to explore the existence of differing clusters within children reported with care and protection concerns, \textbf{3)} to assess the performance of predictive models such as LASSO logistic regression on each subgroup of children (clusters) and, \textbf{4)} to determine whether separate models must be developed for children with specific features to attain better results.   
\section{Methods}
\subsection{ Data}
Our study relied on a unique de-identified data set that we constructed by record linkage between administrative data sets owned by the NZ Ministry for Children (Oranga Tamariki)\footnote{\textbf{Oranga Tamariki}, also known as the \textbf{Ministry for Children}, is a government department in NZ responsible for the well-being of children, specifically children at risk of harm, youth offenders, and also young people that are likely to offend.}, Ministry of Social Development (MSD)\footnote{\textbf{Ministry of Social Development (MSD)} is the lead agency for the NZ social sector that provide employment support, income support, entitlements and superannuation services.}, as well as the NZ 2018 Census records, which are available through the Integrated Data Infrastructure (IDI) managed by Statistics NZ (Stats NZ). There is an encrypted unique identifier (UID) for each identity in the IDI that is common across all data sets. This provides researchers with an ability to link variables from multiple sources to gain system-wide insights. 

Starting with intake records from NZ child protection system (Oranga Tamariki), our research data set included 82,338 notifications involving 55,287 unique children and young people. For inclusion, the notification must have been made in 2019 and under Section 15 of Oranga Tamariki Act 1989. According to Section 15 of the act, any person who believes that a child or young person has been, or is likely to be, harmed, ill-treated, abused, (whether physically, emotionally, or sexually), neglected, or deprived, or who has concerns about the well-being of a child or young person, may report the matter to the chief executive or a constable. These notifications are referred to as care and protection notifications. Since some children and young people are reported more than once during the year or in their lifetime, their initial care and protection notification in 2019 were considered for analysis. Children involved in these notifications were then linked to their records from the NZ public benefit system (MSD), the NZ census and the Stats NZ demographic records to define the outcome variable and predictor variables. A more detailed explanation of the linking process is in Figure 1. 
\begin{figure}[h]
	\centering
		\includegraphics[width=15cm,scale=1]{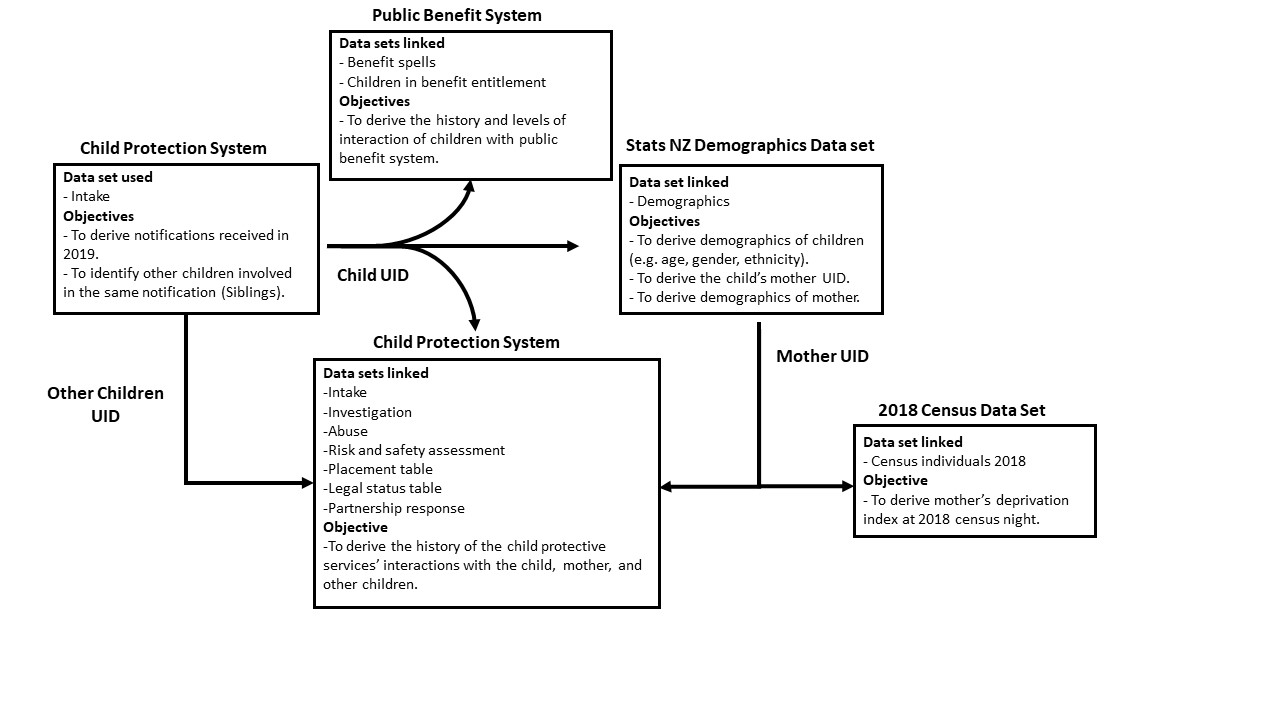}
	  \caption{The process of data linkage and research data set development.}\label{fig1}
\end{figure}
\subsubsection{Outcome variable}
The outcome variable was defined based on events from Table 1. Our model intends to predict, for each report of concern, the probability that one or more of the events in Table 1 would take place within the next two years.
\begin{table}[htb]
\begin{center}
\caption{Care and protection related events.}
\begin{tabular}{p{2cm} p{12cm}}
 \hline
 \textbf{Event 1}&A \textit{substantiated finding of maltreatment}\footnotemark including
 physical, sexual, and emotional abuse or neglect. \\ 
 \hline
 \textbf{Event 2}&A site social worker recommending a 
\textit{Family Group Conference (FGC)}\footnotemark or creating a
 \textit{Family/Whānau Agreement (FWA)}\footnotemark.\\
 \hline
 \textbf{Event 3}&The child or young person being the
 subject to a \textit{further notification} which will be 
assessed as an intake. \\ 
 \hline
\end{tabular}
\end{center}
\end{table}
\footnotetext[4]{\textbf{Substantiated finding of maltreatment} means that the social worker has clear and sufficient evidence to determine that maltreatment has occurred.}
\footnotetext[5]{\textbf{Family Group Conference (FGC)} is a formal meeting where Oranga Tamariki and the extended family of children or young people work together to develop a plan to address any care and protection concerns, needs or wellbeing issues relating to the child or young person.}
\footnotetext[6]{\textbf{Family/Whānau Agreement (FWA)} is an agreement between the family and Oranga Tamariki and deals with concerns for their children and how Oranga Tamariki will support them.}\newline
To allow for two years of records’ follow up, we considered unique children and young people who were under 16 years of age at the time of notification in our analysis. The reason is that child protection data sets often contain records for children and young people under 18 years of age. As a result, the final data set included observations involving 55,287 unique children and young people under the age of 16. From the children and young people notified in 2019, 48 percent were found to have experienced at least one of the events mentioned above within two years indicating that the data is balanced.
\subsubsection{Predictor variables}
The predictor variables can be classified into four main groups such as \textit{child} predictors, \textit{caregiver (mother)} predictors, \textit{family} predictors and \textit{others}. These variables were created using the records available in the data sets we had access to through IDI and based on the risk factors of abuse and maltreatment identified in literate and academic studies, mainly the ones described in \citep{rea2017report}. The list of these variables and their description is represented in the Appendix.
\subsubsection{Handling missing data}
For training the predictive model for the total population, the missing values for the mother's \textit{age} was imputed by the mean. However, the missing age for observations within each cluster was imputed based on the mean age of mothers within that cluster. To deal with the missing values for the categorical variables (e.g.,\textit{gender, Ethnic group, Deprivation Index, etc.}), an additional category indicating the missing value was created (e.g. unknown). More information regarding the categories included in these variables can be found in the Appendix.
\subsection{Clustering analysis }
Clustering is an unsupervised machine learning technique used for grouping data into clusters based on their similarity. In CA process, observations are grouped according to their homogeneity and distinctiveness, regardless of the correlation structure of the predictors \citep{tryfos1998methods}. In practice, it is generally challenging to determine whether clusters are an indication of phenotypic grouping, or if they are simply the result of dependency between variables. Machine learning-wise, we aim to identify reliable clusters of children reported with care and protection concerns using numerical predictor variables presented in Table 2 and then assess our predictive risk models trained on these clusters.


\begin{table}[htb]
\centering
\caption{Numerical predictor variables used for clustering analysis (SD stands for standard deviation).}
\begin{tabularx}{\linewidth}{X c c} 
 \hline
 \textbf{Numerical Variable (n=55,287)} & \textbf{Mean} & \textbf{SD} \\ 
 \hline
 Child age at the time of notification&7.20&4.68\\
 Number of previous care and protection notifications&3.22&4.31 \\ 
Number of days since last intake&609.10&917.01\\
Number of substantiated findings of maltreatment&1.15&2.27\\
Mother’s age at the time of notification&33.70&7.48\\
Number of other children reported at the same time&1.76&1.55\\
Number of previous notifications for the children reported at the same time&7.46&13.16\\
 \hline
\end{tabularx}
\end{table}
\subsubsection{K-Means clustering and Principal Component Analysis}
The K-Means clustering algorithm is a partition-based CA approach where ‘K’ observations are selected as initial cluster centers. Each observation is then assigned to its nearest cluster based on its Euclidean distance to each cluster center. Afterward, all cluster averages are updated, and the process is repeated until a convergence of the criterion function has been achieved \citep{vora2013survey}. The K-Means algorithm clusters large data sets into a specified number of clusters (K) by minimizing the squared error function. Therefore, some data may be misclassified as a result of outliers \citep{prabhu2011improving}. Principal Component Analysis (PCA) has been proven to be a continuous solution to the cluster membership indicators for K-Means clustering \citep{ding2004k,ding2004principal}. PCA is a widely used statistical approach adopted as an effective method for unsupervised dimension reduction by extracting relevant information from data sets  \citep{joliffe1992principal}. It also facilitates the discovery of hidden relationships and  improve data visualization, detection of outliers, and classification within the newly defined dimensions \citep{prabhu2011improving}. \citet{ding2004principal} showed that the continuous solutions of the discrete K-Means clustering membership indicators are the data projections on the principal directions which are the principal eigenvectors of the covariance matrix. PCA reduces the data set to a lower dimension, while ensuring that the least information is lost, and provides a better centroid point for clustering \citep{zhu2019improved}.
\subsubsection{Results}
In this study, PCA was applied before K-Means clustering was performed. As PCA generates a feature subspace that maximizes the variance along the axes, the data set was first standardized onto a unique scale with a Mean of 0 and a Variance of 1. Scaling improves the results produced by PCA which is a requirement for the optimal performance of many machine learning algorithms. Based on the output produced from PCA, and the eigenvalue criterion \citep{boehmke2019hands}, three principal components were selected. After the selection of principal components using this criterion, the components were passed for K-Means clustering. 

The initial step in the application of K-Means clustering algorithm involves the identification of an optimal number of clusters (K). For this purpose, the elbow method was used. Basically, the K-Means clustering algorithm is applied multiple times with different K values, and the within cluster sum of squares is calculated and plotted for each K value (Figure 2).
\begin{figure}[h!]
	\centering
		\includegraphics[width=11cm,scale=2]{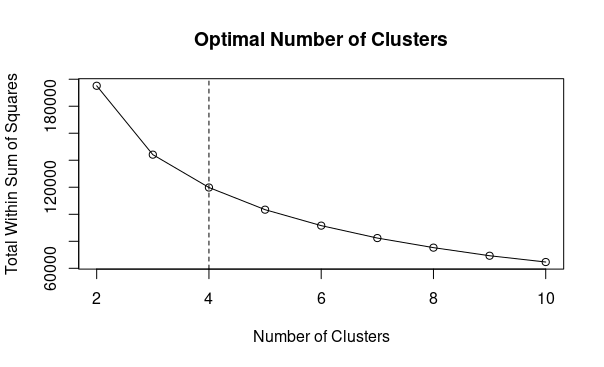}
	  \caption{Elbow plot to select the optimal number of clusters.}\label{fig2}
\end{figure}
\newline On the elbow plot, the optimal number of clusters is defined as the point beyond which there is only a minor reduction in within-cluster variability. This is visually represented as the bend of the elbow. According to our elbow plot (Figure 2), the optimal number of clusters (group of children and young people) should be 4. Consequently, K-Means clustering algorithm (K=4) was applied on the three selected principal components as input. Figure 3 Visualizes the clusters identified by applying K-Means clustering via PCA with principal components (PC1-PC3) on the axes.
\newpage\begin{figure}[h!]
	\centering
		\includegraphics[width=11cm,height=6cm,scale=2]{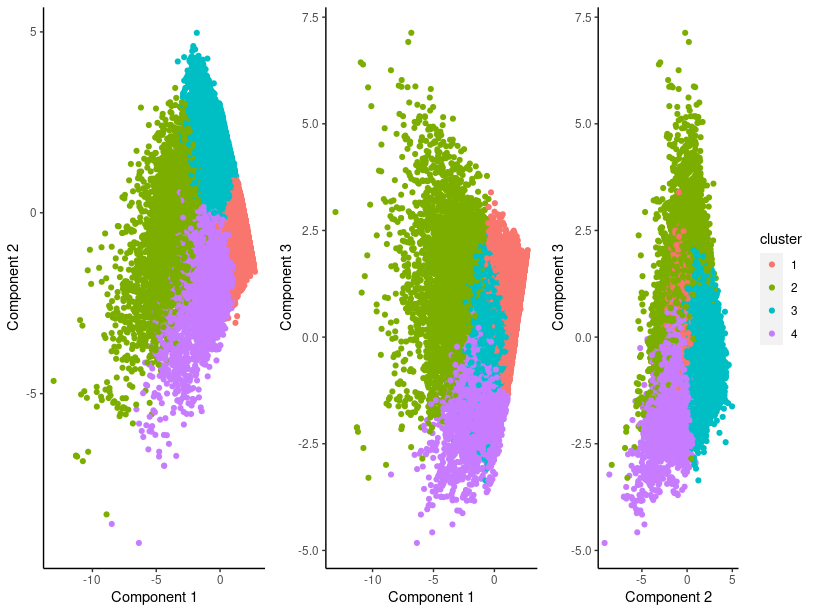}
	  \caption{K-Means clusters (4) derived from PCA.}\label{fig3}
\end{figure}
After completing the CA process and assigning each child to a cluster, we divided the children into four groups (data sets) based on their cluster membership. As a next step, separate models were trained for the total population and each subgroup of children. We describe the modelling process and the results from assessing the performance of trained models in the following section.\
\subsection{ Modelling}
Children and young people notified in 2019, and the data sets created based on their cluster membership were randomly split into a training data set containing 70 percent of records and a testing data set containing the remaining 30 percent. A training data set was used to develop a model, and a testing data set to assess how well the model can correctly identify children who will be subject to care and protection concerns within the next two years. As the main method of modelling in this study, we used LASSO logistic regression.
\subsubsection{LASSO logistic regression}
Logistic regression is a supervised learning classification algorithm used to predict the probability of an outcome variable based on predictor variables. Due to its simplicity and easy interpretation, this statistical modelling algorithm has been widely used in several fields, including biological sciences, social sciences, and machine learning \citep{james2013introduction}. In its simple form, logistic regression utilizes weights for all predictors despite their significance and the potential for model over-fitting. By contrast, the LASSO regularized form of logistic regression effectively selects only the most important predictor variables by shrinking the regression coefficients with the least important predictor variables to zero while minimizing prediction error, given the sum of the absolute value of the weights is less than a constant. Thus, it is capable of both predictor selection and regularization, which results in more easily interpretable and more accurate models \citep{tibshirani1996regression}. As well as these advantages, there were other reasons why we chose LASSO as our method for training models. As a first point, although logistic regression was selected as the most appropriate model in previous NZ studies \citep{wilson2015predictive,vaithianathan2013children,rea2017report}, and version 1 of the AFST, \citep{vaithianathan2017developing}, however LASSO regularized form of logistic regression has not  been tested for the NZ population. Furthermore, in most recent US studies, LASSO was selected as the best model based on its overall performance and accuracy for the specific high-risk groups in addition to equivalent level of accuracy for black children versus non-black children \citep{vaithianathan2019allegheny,vaithianathan2019implementing}. Upon this approach we will be able to compare the results in terms of accuracy with the state-of-the-art approaches in child welfare settings.

The LASSO model was implemented with the R package named glmnet \citep{friedman2021package}. In the model training process, the constant – often symbolized as lambda – was optimized using 10-fold cross validation approach.
\subsubsection{Model performance assessment}
A number of test set classification metrics are provided in Table 3 for the LASSO models trained for the entire population of unique children and young people notified in 2019 and their four clusters. AUC was used to evaluate the performance of LASSO when predicting outcomes using the test data sets. The ROC curves for LASSO applied to four clusters of children are shown in Figure 4.
\begin{table}[htb]
\centering
\caption{Performance results for models under analysis.}
\begin{tabularx}{\linewidth}{X c c c} 
 \hline
 \textbf{Model}&\textbf{AUC}&\textbf{TPR}\footnotemark&\textbf{TNR}\footnotemark\\ 
 \hline
 Entire population&0.68&0.64&0.63\\
Cluster 1&0.70&0.63&0.66 \\ 
Cluster 2&0.63&0.62&0.60 \\ 
 Cluster 3&0.63&0.57&0.61 \\ 
 Cluster 4&0.68&0.66&0.62 \\ 
 \hline
\end{tabularx}
\end{table}
\footnotetext[7]{\textbf{True Positive Rate (TPR)} is the fraction of positive cases correctly predicted to be in the positive class out of all actual positive cases, TP/(TP+FN).}
\footnotetext[8]{\textbf{True Negative Rate (TNR)} is the fraction of negative cases correctly predicted to be in the negative class out of all actual negative cases, TN/(TN+FP).}
\begin{figure}[h!]
	\centering
		\includegraphics[width=11cm,scale=2]{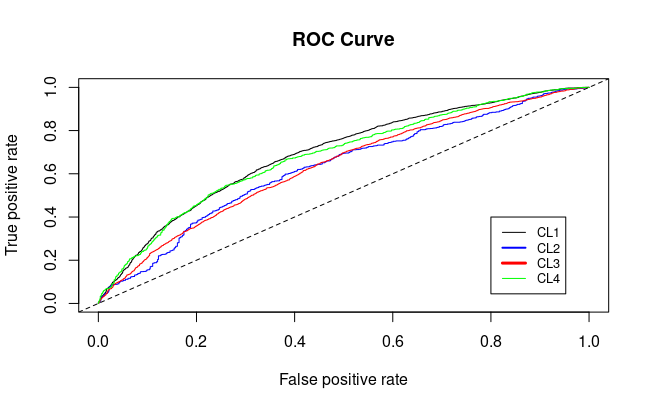}
	  \caption{The ROC curves for LASSO Logistic Regression applied to four clusters of children.}\label{fig4}
\end{figure}\newline
The reason for considering AUC to measure the performance of these models is that child protection systems are often interested in developing a predictive tool that can supplement and standardize clinical decisions through a risk score or a summary statistics weighting information from the administrative data \citep{vaithianathan2017developing}. A tool which allows for the use of empirically derived scores in combination with clinical judgement and other sources of data that are not instantly available to generate a screening decision. In this context, the AUC is a useful statistic for assessing the goodness of fit or prediction accuracy. There are various interpretations of AUC, but the one that is particularly useful in this context is that it can be understood as the probability that a randomly selected child that is a true positive (i.e., has had a care and protection concern) has a higher risk score than a randomly selected child that is a true negative (i.e., has not been the subject of a care and protection concern within 2 years).
\subsubsection{Results}
According to the AUC measures in Table 3 and the ROC curves shown in Figure 4, LASSO appears to perform slightly better for group of children in cluster 1 than other clusters (0.70), whereas its performance is the poorest for group of children in cluster 2 and cluster 3 (0.63). The AUC measures also indicate that the models generated for group of children in cluster 1 and cluster 4 perform very closely. True positive rate (TPR), however, is higher by 3 percent for cluster 4 (0.66), and True Negative Rate (TNR) is higher by 4 percent for cluster 1 (0.66). It is unclear why models trained on subgroups of children in cluster 1 and 4 have slightly lower error rate. However, one way to investigate this was to examine the characteristics of children within each cluster by analyzing descriptive statistics of the variables. 

The baseline characteristics for four clusters are described using standard descriptive statistics for the seven numerical variables used in the cluster analysis (Table 4). In addition to table 4, Figure 5 provide a better representation of cluster characteristics based on these 7 numerical variables.
\begin{figure}[h!]
	\centering
		\includegraphics[width=14cm,height=11.5cm,scale=2]{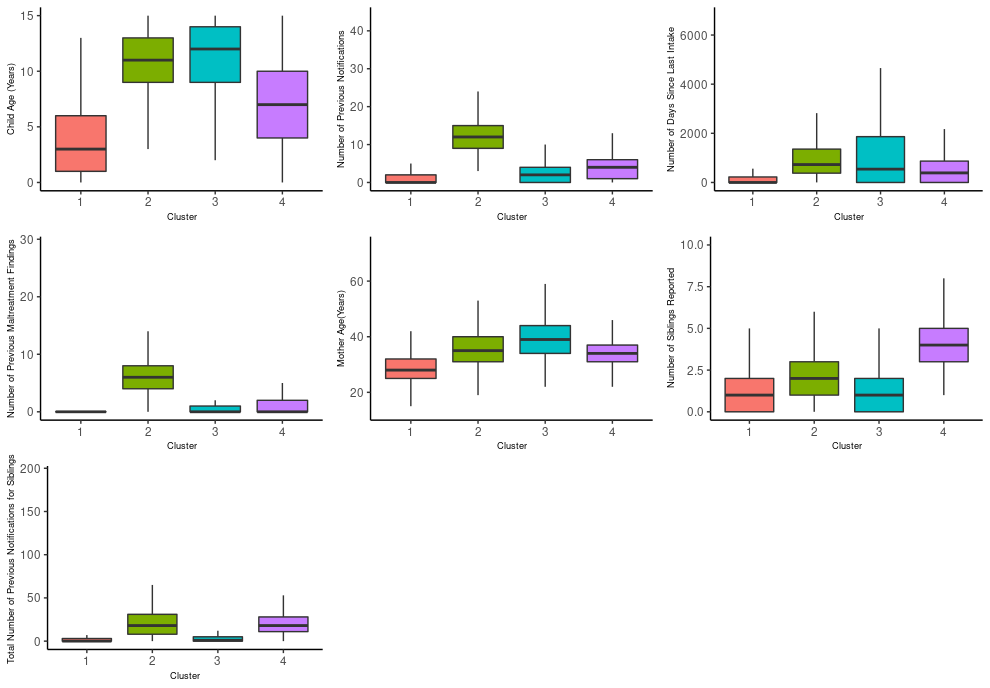}
	  \caption{Distribution of numerical variables in four clusters.}\label{fig5}
\end{figure}\newpage
\begin{table}[htb]
\centering
\caption{Descriptive statistics for four clusters (data in the table are Means ± SD).}
\begin{tabularx}{\linewidth}{X c c c c} 
 \hline
 \textbf{Variables}\footnotemark &\textbf{Cluster1}&\textbf{Cluster2}&\textbf{Cluster3}&\textbf{ Cluster4}\\ 
  &\textbf{(n=23,658)}&\textbf{(n=5,544)}&\textbf{(n=17,901)}&\textbf{(n=8,187)}\\ 
 \hline
Var1 &3.44 ± 2.94&10.78 ± 2.96&11.18 ± 2.86&6.92 ± 3.85\\
 Var2 & 1.11 ± 1.75&12.50 ± 5.02& 2.81 ± 2.89&3.91 ± 3.03\\ 
Var3 & 193.90 ± 397.11&979 ± 805.02&1050 ± 1233.32&594.10 ± 665.15\\
 Var4 & 0.30 ± 0.84&6.21 ± 3.38&0.70 ± 1.17&1.15 ± 1.47\\
Var5&28.41 ± 5.29&35.47 ± 6.12&39.47 ± 6.49&34.10 ± 4.91\\
Var6 &1.25 ± 1.04&2.14 ± 1.56&1.24 ± 1.08&4.11 ± 1.36\\
Var7 & 2.31 ± 3.83&22.56 ± 20.85&2.96 ± 4.23&21.96 ± 17.58\\
 \hline
\end{tabularx}
\end{table}
\footnotetext[9]{Var1:= Child age at the time of notification ; Var2:=Number of previous care and protection notifications; Var3:=Number of days since last intake; Var4:= Number of substantiated findings of maltreatment; Var5:= Mother’s age at the time of notification; Var6:=Number of Other children reported at the same time; Var7:= Number of previous notifications for the children reported at the same time.}
Based on a comparison of the means within each cluster (Table 4), and the distribution of numerical variables illustrated by Figure 5, it appears that the means are lower for children in cluster 1. Specifically, the average age is lower for group of children in cluster 1 and cluster 4 with a higher AUC. As a result, there is a possibility that models will perform differently for young children as opposed to older children. To investigate this further, we categorized the children under 16 years of age notified in 2019 into different age groups and trained the Lasso models separately on each group. Several test set classification metrics are provided in Table 5.

\begin{table}[htb]
\centering
\caption{Performance results for models developed based on children age group.}
\begin{tabularx}{\linewidth}{X c c c} 
 \hline
 \textbf{Age group}&\textbf{AUC}&\textbf{TPR}&\textbf{TNR}\\ 
 \hline
Newborn&0.71&0.67&0.64\\
 Newborn to Five&0.70&0.66&0.62\\ 
 Newborn to Ten&0.68&0.66&0.61\\ 
 Newborn to 15&0.68&0.64&0.63\\ 
 Six to Ten&0.68&0.63&0.63\\ 
Eleven to Fifteen&0.67&0.60&0.64\\
 \hline
\end{tabularx}
\end{table}

Although the results did not demonstrate a significant difference in AUC for different age groups, the models did perform slightly better for children under the age of 5, and especially for new-born children, TPR and TNR appeared to be the highest (Table 5). 
\section{Discussion and future work}
This study was conducted based on the data from 55,287 unique children and young people aged under 16 who were reported to the NZ child protective services in 2019. During the predictive risk modeling process using our developed research data set, the LASSO algorithm trained on the entire population didn’t perform as expected to produce a model to the desired predictive ability. While this may have been a consequence of the limited number of predictors, but we examined whether certain characteristics of children and young people were also influencing its performance. This paper presented our initial findings of our investigation on the identification of such features through CA techniques, as well as their potential effect on risk-modelling.

In a novel approach not previously considered in other studies from NZ, we utilized K-Means clustering via PCA to identify subgroups (clusters) of children based on observations of the numerical variables presented in Table 2. As well as being an effective method for reducing dimensions and extracting relevant data from data sets without supervision, PCA has been shown to facilitate the discovery of hidden relationships, improve data visualization, and detect outliers, resulting in more effective K-Means clustering \citep{prabhu2011improving,ding2004k,ding2004principal}. From the CA via PCA, children were divided into four groups and separate models were trained using LASSO. Results from CA and predictive modeling process revealed no significant differences in their performance (Table 3). Two models, however, performed slightly better (Cluster 1 and Cluster 4) including younger children only (see Table 3 and Table 4). This is further evidence that our research data set may not capture all potential risk factors for older children or generally predictive models are more accurate when developed for children under the age of five (Table 5). These assumptions are consistent with \cite{palusci2011risk} who examined risk factors for child maltreatment among US children. Infants and young children in the child welfare system were found to have different risk factors and are provided with different services than older children. Results from our study strongly suggest that age is a crucial factor for predicting child maltreatment and we are currently investigating this matter. 

Generally, the predictive risk models trained in this study did show a slightly lower-than-standard predictive power and we could not find a significant difference in performance between the models trained for each cluster. However, our results can help identify potential factors to consider in future stages. Given that our research data set may not capture all potential risk factors or predictors for older children, further research is required to determine whether the addition of new predictors will contribute to accuracy improvements of the models and reduction of the error rates (see Table 3). Consider, for example, the data set created here included variables related to the mother (see Appendix), but similar variables could be generated for the father as well. Furthermore, it is possible to create new predictors by combining data from other organizations. Records from the Department of Corrections available in the Stats NZ IDI database may also be extracted to verify whether a child lives with and adults who has recently been released from prison for an offence related to family violence. A further source of predictors is data provided by the Ministry of Health concerning addictions and mental health records of caregivers of children. Considering that these characteristics are identified as risk factors for child maltreatment \citep{ayers2019perinatal,lopes2021parental,austin2016prior}, incorporating predictor variables that reflect these factors might enhance the accuracy of models that are developed for predicting child maltreatment.

Furthermore, our results on younger children, involving more accurate models, shed light on the importance of such age group in this analysis and can be used by child protection systems in other situations. In this work we investigate the modeling framework for potential use when notification is received by NZ child protective services and for children and young people under the age of 16. But, there are other potential situations in which a predictive risk model can be utilized, including pre or immediately after the childbirth. Moreover, using the model pre or post childbirth help to identify children who are at risk and provide high risk families with appropriate services in advance. For such models we will need to add live events data and maternity data available in Stats NZ IDI database. Maternity data helps gain information about the baby’s condition at birth and mother’s condition before and after giving birth to the child. Life event data is about life events relating to births, deaths, marriages, and civil unions registered in NZ. 

Aside from the absence of certain predictors, there are other factors that may affect the performance of predictive risk models. Training models on biased data is one of them. Crucially, the source of any predictive model is the data on which the algorithm is trained. Any source of error will be translated into the output of the model \citep{barocas2016big}. In particular, an error that changes an important factor about a child will reduce the accuracy of the model for that child \citep{glaberson2019coding}. In the context of child welfare predictive tools, there are reasons to be concerned about the level of error present in the data being fed into the algorithm \citep{glaberson2019coding,rea2017report}. The data used to develop such predictive models are often extracted from the child welfare agency’s database systems and are linked to data collected by other government agencies.  Data from government administrative systems is entered, initially, by human and therefore is subject to human error. For instance, names, addresses, or other vital information may be incorrect, information from one individual may be incorrectly linked to another, old and outdated information may persist, or information may be missing altogether \citep{glaberson2019coding,rea2017report}. Feasibility studies and ethical reviews on the use of these models suggest that although the linkage of administrative data to support predictive risk modeling is feasible, but the linkage is subject to error and a system for review would be needed in any implementation \citep{ministry2014feasibility}. This work, on the other hand, rely upon Stats NZ’s data quality and have used identifiers created by Stats NZ for data linking. 

Moreover, an important ethical concern has been whether predictive analytics methods will worsen existing racial disparities in child protection systems. In particular, past studies suggest that the presence of persistent racial bias reflected in administrative data have the potential to increase error rates and might lead to discrimination or unfairness against a certain group of people \citep{cuccaro2017risk}. This is  an important factor that needs to be addressed during the process of developing predictive risk models. Specifically in the NZ child welfare system, the over-representation of indigenous people of NZ (Māori) or other low socioeconomic status groups in the child welfare systems might be intensified by predictive risk models. If the data exaggerates risk, then its use in decision-making has the potential to feed a cycle of bias that leads to different population groups (such as Māori) being disadvantaged or discriminated against \citep{rea2017report}. Further work will be required to explore these findings and make sure the model can identify the risk as accurately as possible and does not unintentionally add to an over-representation of Māori within the NZ child welfare system.  

\newpage

\bibliography{./References.bib} 
\bibliographystyle{apacite} 

\newpage
\appendix
\begin{landscape}
\section*{Appendix}
\subsection*{Table A1. Child predictors.}
\centering
\begin{tabular}{p{8cm}p{2.5cm}p{3.5cm}p{8.5cm}}
    \hline
    \textbf{Variable} &\textbf{Type} & \textbf{Coding Definition} & \textbf{Description} \\
    \hline
    Age(Years) & Numerical & Integer number &  Age at the time of notification.\\ 
    \hline
    Gender & Categorical&Male\newline Female \newline Unknown& Gender of the child or young person.\\
    \hline
    Ethnic group & Categorical & Māori\newline Māori and Pacific \newline Pacific\newline European\newline Other\newline Unknown & Since one child can have more than one ethnicity, this variable was created based on OT\footnotemark approach regarding prioritised ethnicity.\begin{itemize}\setlength\itemsep{-0.5em}
 \item Māori\footnotemark children who identify Māori (but not Pacific) as one of their ethnicity.
 \item Māori and Pacific children who identify both Māori and Pacific as their ethnicity.
 \item Pacific children who identify Pacific (but not Māori) as one of their ethnicity.
 \item New Zealand European and Other children who do not identify Māori or Pacific as one of their ethnicity.
\end{itemize}\\
\hline
Previous risk and safety assessment flag & Binary & 1,0 & This variable indicates whether the child has previously been the subject of a risk and safety assessment.\\
\hline
Number of previous care and protection notifications & Numerical & Integer value & This variable includes the number of previous care and protection notifications for the child.\\
\hline   
 No previous care and protection notification flag & Binary & 1,0 & Since the above variable is zero inflated, this binary variable was created to indicate whether the child has not been the subject of a care and protection notification in the past.\\
 \hline
   \end{tabular}
  \footnotetext[1]{The Māori are the indigenous Polynesian people of mainland New Zealand (NZ). Māori originated with settlers from East Polynesia, who arrived in NZ between roughly 1320 and 1350.}
   \footnotetext[2]{OT stands for Oranga Tamariki or NZ Ministry for Children. It is a NZ government agency that is responsible for the well-being of children, specifically children at risk of harm, youth offenders, and also young people that are likely to offend. }
 \newpage
  \textit{Table A1 Continued.}\vspace{0.25cm}
   \centering
\begin{tabular}{p{8cm}p{2.5cm}p{3.5cm}p{8.5cm}}
    \hline
    \textbf{Variable} &\textbf{Type} & \textbf{Coding Definition} & \textbf{Description} \\
    \hline
 Number of days since last intake & Numerical & Integer value & Number of days since the child was the subject of a Section 15 intake where further action was required by child protection services. This is not including the current notification.\\ 
\hline 
No previous intake flag & Binary & 1,0 & This variable indicates whether the child has not been the subject of a notification with an intake outcome in the past.\\
\hline
Number of previous maltreatment findings & Numerical & Integer value & This variable includes total number of previous substantiated findings of maltreatment for the child including emotional abuse, physical abuse, sexual abuse, and neglect.\\
\hline
No previous maltreatment finding & Binary & 1,0 & This variable indicates whether the child has not been the subject of a maltreatment finding in the past.\\
\hline
Previous custody guardianship spell flag & Binary & 1,0 & This variable indicates whether the child has previously had care and protection custody or guardianship to the Chief Executive of OT or another service provider.
\newline This does not include section 205 which is a temporary order and section 42 which allows a police constable, who believes its necessary to protect a child from injury or death and detain the child.\\
\hline
Open phase flag & Binary & 1,0 & This variable indicates whether the child is already in an open social work phase at the time of notification such as:\begin{itemize}\setlength\itemsep{-0.5em}
 \item Investigation
 \item Risk and safety assessment
 \item Partnered response
 \item Placement
\end{itemize}\\
\hline
  \end{tabular}
 \newpage
  \textit{Table A1 Continued.}\vspace{0.25cm}
   \centering
\begin{tabular}{p{8cm}p{2.5cm}p{3.5cm}p{8.5cm}}
    \hline
    \textbf{Variable} &\textbf{Type} & \textbf{Coding Definition} & \textbf{Description} \\
    \hline
Main public benefit inclusion flag & Binary & 1,0 & This variable indicate whether the child is included in a care giver’s main benefit currently or in the past.
The main benefit here refers to:
\begin{itemize} \setlength\itemsep{-0.5em}
 \item Sole parents 
 \item Job seekers 
 \item Support living payment 
 \item Young parent 
\end{itemize}\\
\hline
Level of contact with MSD\footnotemark and OT & Ordinal & 1,2,3,4 &
Two binary variable were created in order to define this variable. A variable that indicated whether the child has previously had a contact with NZ public benefit system (MSD) and another one that indicated whether the child has previously been in contact with NZ child protective services (OT). For the final variable which is used in the analysis, each child is categorised into one of four groups:
 \newline\textbf{Level 1}: No previous public benefit system or OT contact.\newline\textbf{Level 2}: Previous OT contact, no previous contact with public benefit system.\newline\textbf{ Level 3}: Previous contact with public benefit system, no previous OT contact. \newline\textbf{Level 4}: public benefit system and OT contact.\newline\\
\hline
 \end{tabular} 
 \footnotetext[3]{MSD stands for the NZ Ministry of Social Development. It is the lead agency for the NZ social sector that provides employment support, income support, entitlements and superannuation services.}\\
\subsection*{Table A2. Care giver predictors.}
\centering
\begin{tabular}{p{8cm}p{2.5cm}p{3.5cm}p{8.5cm}}
    \hline
  \textbf{Variable} &\textbf{Type} & \textbf{Coding Definition} & \textbf{Description} \\
    \hline
    Age (Years) & Numerical & Integer value & Age of the child's mother at the time of notification.\\
\hline
Level of contact with child protective services & Ordinal & 1,2,3,4&
The level of contact with child protective services during the childhood of the care giver (mother).\newline\textbf{Level 1}: No involvement. \newline\textbf{Level 2}: At least one intake as a child.\newline\textbf{ Level 3}: Finding of maltreatment. \newline\textbf{Level 4}: Placement.
\newline\textbf {Note}: We are aware that complete history of contact with child protective services is only available for younger care givers.
\newline\\
\hline
NZ deprivation index & Categorical & 1 ,2, 3, 4\newline 5, 6, 7, 8\newline 9, 10, unknown & NZ Deprivation Index for the care giver based on 2018 census.\\
\hline
\end{tabular}
\vspace{2cm}
\subsection*{Table A3. Family predictors.}
\centering
\begin{tabular}{p{8cm}p{2.5cm}p{3.5cm}p{8.5cm}}
\hline
 \textbf{Variable} &\textbf{Type} & \textbf{Coding Definition} & \textbf{Description}\\
 \hline
 \raggedright Number of children reported at the same time &	Numerical & Integer value & The number of children involved in the notification (siblings).\\
 \hline
 Number of previous notifications for children reported & Numerical & Integer value & Total number of previous notifications for the children involved in the notification (siblings).\\
 \hline
 \end{tabular}
\subsection*{Table A4. Other predictors.}
\centering
\begin{tabular}{p{8cm}p{2.5cm}p{3.5cm}p{8.5cm}}
\hline
 \textbf{Variable} & \textbf{Type} & \textbf{Coding Definition} & \textbf{Description} \\
 \hline
 Notifier’s role & Categorical &\raggedright -Anonymous\newline -Court \newline -Family \newline -Health Professionals \newline -Midwife or Plunket\footnotemark \newline -Neighbours or Friends \newline -Police (FVI\footnotemark) \newline -Police (Other)\newline -School or Early Childhood Centre\newline -Unknown \newline -Others  \newline & This variable is aggregated and includes the role of the notifier.\\
 \hline
 \end{tabular}
  \footnotetext[4]{Plunket is a national charitable organisation and is NZ's largest provider of support services for the development, health and wellbeing of children and families.}
 \footnotetext[5]{The police notifies child protection agency after attending a family violence incident where a child is present.}
\end{landscape}

\end{document}